\documentclass[runningheads]{llncs}
\usepackage{graphicx}
\usepackage{amsmath}
\newcommand{\betP}{\mathrm{BetP}}

\newcommand{\odis}{\mathrm{oDis}}
\newcommand{\oDP}{\mathrm{oDP}}
\newcommand{\mix}{\mathrm{Mix}}

\newcommand{\bel}{\mathrm{bel}}
\newcommand{\pl}{\mathrm{pl}}
\newcommand{\argmax}{\operatornamewithlimits{argmax}}

\newcommand{\Conf}{\mathrm{Conf}}

\newcommand{\Ocup}{\operatornamewithlimits{\text{$\stackrel{o}{\cup}$}}}

\begin{document}
\title{Belief functions on ordered frames of discernment}
%
%
\author{Arnaud Martin}
\authorrunning{A. Martin}
%
\institute{Univ Rennes, CNS, IRISA, DRUID, IUT de Lannion, France
\email{arnaud.martin@irisa.fr}\\
\url{http://people.irisa.fr/Arnaud.Martin}}
\maketitle              
\begin{abstract}
Most questionnaires offer ordered responses whose order is poorly studied via belief functions. In this paper, we study the consequences of a frame of discernment consisting of ordered elements on belief functions. This leads us to redefine the power space and the union of ordered elements for the disjunctive combination. We also study distances on ordered elements and their use. In particular, from a membership function, we redefine the cardinality of the intersection of ordered elements, considering them fuzzy. 

\keywords{ordinal variable \and ordered frame of discernment  \and ordered and fuzzy elements \and ordered power set \and distance.}
\end{abstract}
\section{Introduction}
The theory of belief functions is used in more and more applications such as machine learning, pattern recognition, clustering, etc. To apply the theory of belief functions, we consider the frame of discernment given by \linebreak $\Omega=\{\omega_1, \omega_2, \ldots, \omega_n\}$ and the basic belief assignments are defined as a mapping from the power set of $\Omega$, noted $2^\Omega$, to $[0,1]$ \cite{Shafer76}. The elements $\omega_i$ of $\Omega$ are considered exclusive and exhaustive. The assumption of the exclusivity can be lifted, considering the hyper power set $D^\Omega$ \cite{Martin09}. 

However, we may be confronted with applications where a semantic or proximity link exists between the elements of the frame of discernment. For example, in questionnaires graduated or ordered answers can be proposed: \\
What is the distance between Lannion and Paris?:\\
\null \quad \textit{314km, 414km, 514km, 614km}\\
Other possible answers use the Likert scale such as:\\
\null \quad \textit{Not happy, neutral, happy}\\
\null \quad \textit{Strongly disagree; Disagree; Neither agree nor disagree; Agree; Strongly agree}

These answers can be represented by an ordinal variable. In these cases, there is no way to deal with this link between the elements of the frame of discernment. Forcing these elements to have a zero mass is not stable in the combination and decision processes. 
In this paper, we explore the possibilities of considering an order between the elements of the frame of discernment with the theory of belief functions. First in section~\ref{oPS} we introduce the ordered power set, then we show how we can combine mass functions defined on an ordered power set in section~\ref{combination}. In section~\ref{distoPS}, we propose the definition of a distance between ordered elements used to define a distance between belief functions, that can be used for conflict measure and decision in section~\ref{decision}. In section~\ref{fuzzyOrdered}, we consider ordered fuzzy elements to redefine the cardinality of the intersection of elements before concluding.

\section{Power set of ordered elements}
\label{oPS}
Let us take again the example of the question about the distance between Paris and Lannion. The frame of discernment is given by $\Omega=\{\omega_1, \omega_2, \omega_3, \omega_4\}$, with the answers $\omega_1=314$km, $\omega_2=414$km, $\omega_3=514$km, $\omega_4=614$km. The right answer is 514km. Here, the elements of the frame of discernment are obviously exclusive. However, if someone answers 614km, the error is smaller than if the answer given is 314km. If we consider imprecise answers as in~\cite{Thierry19}, $\{\omega_3,\omega_4\}$ make sense while $\{\omega_1,\omega_3\}$ does not. 

We therefore consider that we should not take into account all the elements of the power set $2^\Omega$, but only those which have a meaning. Thus disjunctions that do not contain consecutive elements should not be considered. Let us consider an ordinal variable having values in a finite set $\Omega=\{\omega_1,\ldots, \omega_n\}$ of ordered $\omega_i$, $i=1,\ldots,n$, $i$ being an ordinal number. 

\textbf{Definition} \textit{The {\em ordered power set}, noted $oPS^\Omega$, is a subset of the power set composed by the empty set and all the disjunctions of consecutive elements of $\Omega$.}

A disjunction of consecutive elements from endpoint elements is noted by:
\begin{equation}
 \{\omega_i,\omega_j\}_o= \{\omega_i,\omega_{i+1},\ldots,\omega_{j-1},\omega_j\}, \, \mbox{with} \, 1\leq i\leq j \leq n
\end{equation}

Hence, the ordered power set is given by:
\begin{equation}
 oPS^\Omega= \{ \emptyset, \left\{ \{\omega_i,\omega_j\}_o \right\}_{i,j=1,\ldots,n}\}
\end{equation}

The number of elements of $2^\Omega$ is $2^n$, but it is smaller for $oPS^\Omega$.

\textbf{Proposition} \textit{The number of elements of $oPS^\Omega$, with: $\Omega=\{\omega_1,\ldots, \omega_n\}$ is: 
\begin{equation}
1+\frac{n(n+1)}{2} 
\end{equation}
}

\textbf{Proof} The set $oPS^\Omega$ contains the empty set and ordered elements determined by the endpoint elements. The number of these elements is the number of pairs $(i,j)$ with $1\leq i\leq j \leq n$. This number is $\frac{n!}{2! (n-2)!}=\frac{n(n+1)}{2}$. So $|oPS^\Omega|=1+\frac{n(n+1)}{2}$.
\hfill $\qed$

Let $\Omega=\{\omega_1,\ldots, \omega_n\}$ be a frame of discernment of ordered, exclusive and exhaustive elements $\omega_i$, $i=1,\ldots,n$. A mass function $m$ is the mapping from elements of the ordered power set $oPS^\Omega$ onto $[0,1]$ such that:
\begin{equation}
\sum_{X\in oPS^\Omega} m(X)=1.
\label{normDST}
\end{equation}
A focal element $X$ is an element of $oPS^\Omega$ such that $m(X)\neq 0$.
 
From this definition of mass function, all the definitions of special mass functions are available (simple mass functions, non dogmatic, consonant, etc.). A categorical mass function with $m(X)=1$ is noted $m_X$. 
Definitions of classical belief functions such as credibility, plausibility and pignistic probability are also available on the ordered power set $oPS^\Omega$. 
The credibility function is given for all $X \in oPS^\Omega$ by:
\begin{eqnarray}
\label{bel}
\bel(X)=\sum_{Y \subseteq X, Y \neq \emptyset} m(Y).
\end{eqnarray}
The plausibility function is given for all $X \in oPS^\Omega$ by:
\begin{eqnarray}
\label{pl}
\pl(X)=\sum_{Y \in  oPS^\Omega, Y\cap X \neq \emptyset} m(Y).
\end{eqnarray}
However, it is not possible to compute this function as a dual of the credibility function, because an ordered power set is not invariant by the complement. Indeed, 
if we consider $\Omega=\{\omega_1,\omega_2,\omega_3\}$, the complementary of $\omega_2$ is $\{\omega_1,\omega_3\} \notin oPS^\Omega$. The pignistic probability~\cite{Smets90} can be written for all $\omega \in \Omega$ by:
\begin{eqnarray}
\label{pignistic}
\betP(\omega)=\sum_{X \in oPS^\Omega, X \neq \emptyset, \omega \in X} \frac{1}{|X|} \frac{m(X)}{1-m(\emptyset)}.
\end{eqnarray}

\section{Combination of belief functions on ordered power set}
\label{combination}
When considering several mass functions on an ordered power set $oPS^\Omega$ from several sources or persons, we need to be able to combine them. The combination operator must therefore produce a mass function in the same ordered power set $oPS^\Omega$. Therefore, an ordered power set must be invariant by the combination operator. There are a large number of combination operators~\cite{Martin19} and not all of them verify this property.

\textbf{Proposition} \textit{An ordered power set is invariant by the conjunctive combinations (normalized, not normalized, Yager~\cite{Martin19}).}

\textbf{Proof} Let two focal elements of an ordered power set $oPS^\Omega$: $\{\omega_{i_1},\omega_{j_1}\}_o$ and $\{\omega_{i_2},\omega_{j_2}\}_o$. The intersection of these elements is empty or given by:\\ 
$$\{\max(\omega_{i_1},\omega_{i_2}),\min(\omega_{j_1},\omega_{j_2})\}_o \in oPS^\Omega$$

Therefore, an ordered power set is invariant by all conjunctive operators.\hfill $\qed$

\textbf{Proposition} \textit{An ordered power set is not invariant by the disjunctive combination.}

\textbf{Proof} Let two categorical mass functions $m_{\omega_1}$ and $m_{\omega_3}$, with two non consecutive focal elements such as $\omega_1$ and $\omega_3$. The disjunctive combination of these two mass functions is the categorical mass function with $\{\omega_1,\omega_3\}$ such as unique focal element. As $\{\omega_1,\omega_3\} \notin oPS^\Omega$, the ordered power set is not invariant by the disjunctive combinations.\hfill $\qed$

Thus, an ordered power set is not invariant by all the mixed combination operators based on the disjunctive operator. Here we propose a new disjunctive combination operator that makes an ordered power set invariant. Let two elements $Y_i$ and $Y_j$ $\in oPS^\Omega$, we note $Y_i=\{\omega_{i_1},\omega_{i_{n_i}}\}_o$ and $Y_j=\{\omega_{j_1},\omega_{j_{n_j}}\}_o$. We define the union of these two ordered elements by:
\begin{equation}
 Y_i \Ocup Y_j=\{\min(\omega_{i_1},\omega_{j_1}),max(\omega_{i_{n_i}},\omega_{j_{n_j}})\}_o
\end{equation}
The union of $s$ ordered elements is given by the extension of the previous equation or recursively by $((Y_1 \Ocup Y_2)\Ocup Y_3)\ldots \Ocup Y_s$.

Let two mass functions defined on the ordered power set $oPS^\Omega$, for all \linebreak $X \in oPS^\Omega$, the disjunctive combination is given by:
\begin{eqnarray}
\label{disjunctiveO2}
m_\odis(X)=\displaystyle\sum_{Y_i \Ocup Y_j = X} m_i(Y_i)m_j(Y_j).
\end{eqnarray}

The disjunction combination of $s$ mass functions on the ordered power set $oPS^\Omega$, is given for all $X \in oPS^\Omega$ by:
\begin{eqnarray}
\label{disjunctive}
m_\odis(X)=\displaystyle\sum_{Y_1 \Ocup \ldots \Ocup Y_s = X}  \prod_{j=1}^S m_j(Y_j).
\end{eqnarray}
where $Y_j \in oPS^\Omega$ is a focal element of the source $S_j$, and $m_j(Y_j)$ the associated mass function. We can thus rewrite the mixed combination operators~\cite{Martin19}, such as the Dubois and Prade one~\cite{Dubois88} given for all $X \in oPS^\Omega$, $X\neq \emptyset$ by: 
\begin{eqnarray}
\label{DP}
m_\oDP(X)=\sum_{Y_1 \cap \ldots \cap Y_s = X} \prod_{j=1}^S m_j(Y_j)+\sum_{
\scriptstyle{\begin{array}{c}
Y_1 \Ocup \ldots \Ocup Y_s = X\\
Y_1 \cap \ldots \cap Y_s = \emptyset \\
\end{array}}} \prod_{j=1}^S m_j(Y_j).
\end{eqnarray}

\textbf{Proposition} \textit{An ordered power set is stable by the average combination.}

\textbf{Proof} The proof is obvious, because the set of focal elements obtained by the combination is the union of the sets focal elements from the mass functions to be combined. \hfill $\qed$

\section{Distances on belief functions on ordered power set}
\label{distoPS}
 
In the context of belief functions, we often use a distance between mass functions, in order to measure similarity for clustering, to define some measures or to decide. There are many distances that can be considered~\cite{Jousselme11}. 
The most commonly used distance between the belief functions is the distance defined in~\cite{Jousselme01}. This distance can obviously be defined for two mass functions $m_1$ and $m_2$ on $oPS^\Omega$ by:
\begin{eqnarray}
\label{dJ}
d_J(m_1,m_2)=\sqrt{\frac{1}{2} (m_1-m_2)^T\underline{\underline{D}}(m_1-m_2)},
\end{eqnarray}
where $\underline{\underline{D}}$ is an $(1+\frac{n(n+1)}{2}) \times (1+\frac{n(n+1)}{2})$ matrix based on Jaccard dissimilarity whose elements are:
\begin{eqnarray}
\label{DMatrix}
D(A,B)=\left\{
\begin{array}{l}
1, \, \mbox{if} \, A= B=\emptyset,\\
\\
\displaystyle \frac{|A\cap B|}{|A \Ocup B|}, \, \forall A, B \in oPS^\Omega.\\
\end{array}
\right.
\end{eqnarray}
where $|X|$ is the cardinality of $X\in oPS^\Omega$. Of course we have: $$\displaystyle\frac{|A\cap B|}{|A \Ocup B|}=\displaystyle\frac{|A\cap B|}{|A \cup B|}$$
because if $A\cap B\neq \emptyset$ then $A \Ocup B = A \cup B$ with $A, B \in oPS^\Omega$.

However, with this distance, we have without any distinction: 
$$d_J(m_{\omega_1},m_{\omega_2})=d_J(m_{\omega_1},m_{\omega_3})=1.$$
The value 1 is the maximum value of this distance.

\subsection{Distance between ordered elements}
Let us take the example of the question about the distance between Paris and Lannion. We have noticed that if someone answers 614km, the error is smaller than if the answer given is 314km. Thus, the distance between elements of an ordered frame of discernment $\Omega$ can be considered differently according to their order. On a Likert scale, we have the same argument. We can consider that the minimal distance between two elements of $\Omega$ is that between two consecutive elements.  
Inthe same way, the maximum distance between two elements of $\Omega$ is the distance between the first $\omega_1$ and the last $\omega_n$ element of $\Omega$. 

In order to have a distance with a normality property, we propose the following distance between two elements of $\Omega$:
\begin{equation}
 d_o(\omega_i,\omega_j)=\frac{|i-j|}{n-1}
\end{equation}
where $|x|$ is the absolute value of $x =1, \ldots, n$. This distance takes obviously its values in $[0,1]$.

The distance between an element of $\Omega$ and an element $X= \{\omega_{1_x},\omega_{n_x}\}_o$ of $oPS^\Omega$ can be defined by one of the following equations:
\begin{equation}
\label{d_min}
 d_{o_{min}}(\omega_i,X)=\min(d_o(\omega_i,\omega_{1_x}),d_o(\omega_i,\omega_{n_x}))
 \end{equation}

\begin{equation}
\label{d_max}
 d_{o_{max}}(\omega_i,X)=\max(d_o(\omega_i,\omega_{1_x}),d_o(\omega_i,\omega_{n_x}))
 \end{equation}

\begin{equation}
\label{d_mean}
 d_{o_{av}}(\omega_i,X)=\frac{1}{[X|}\sum_{k=1_x}^{n_x}d_o(\omega_i,\omega_{k})
 \end{equation}
 
 The distance between two elements $X= \{\omega_{1_x},\omega_{n_x}\}_o$ and $Y= \{\omega_{1_y},\omega_{n_y}\}_o$ of $oPS^\Omega$ can be defined by one of the following equations:
\begin{equation}
 d_{o_{min}}(X,Y)=\min_{\omega_{y_i}\in Y} d_{o_{min}}(X,\omega_{y_i})
 \end{equation}
\begin{equation}
 d_{o_{max}}(X,Y)=\max_{\omega_{y_i}\in Y} d_{o_{max}}(X,\omega_{y_i})
 \end{equation}

\begin{equation}
\label{dMean}
 d_{o_{av}}(X,Y)=\frac{1}{|XY|}\sum_{k_x=1_x}^{n_x}\sum_{k_y=1_y}^{n_y}d_o(\omega_{k_x},\omega_{k_y})
 \end{equation}
 These distances take their values in $[0,1]$. The use of one of these distances rather than another may depend on the application. For the sake of simplicity, we will note $d_o$ in the following.

\subsection{Distance between belief functions}
As we have seen, the Jousselme distance does not fit the ordered elements of the frame of discernment. Indeed, with Jaccard dissimilarity, the dissimilarity is zero if the intersection is empty. However, on ordered and exclusive elements, the dissimilarity can be different depending on the order. Therefore, we modify the dissimilarity of Jaccard on empty intersection elements from the distance defined in the previous section. Since the minimum strictly positive value of the Jaccard dissimilarity is $\frac{1}{n}$, the proposed dissimilarity takes its values on $[0,\frac{1}{n}]$. Thus, we define a modified Jaccard dissimilarity for ordered elements by:
\begin{eqnarray}
\label{Do}
D_o(A,B)=\left\{
\begin{array}{l}
1, \, \mbox{if} \, A= B=\emptyset,\\
\\
\displaystyle \frac{|A\cap B|}{|A \Ocup B|}+(1-Int(A,B))\frac{1-d_o(A,B)}{n}, \, \forall A, B \in oPS^\Omega.\\
\end{array}
\right.
\end{eqnarray}
where $Int$ is the intersection index defined such as in~\cite{Jousselme11} by $Int(A, B) = 1$ if $A\cap B \neq \emptyset$ and 0 otherwise.

According to~\cite{Bouchard13}, $D_o$ define a matrix positive definite. The distance obtained is given for two mass functions $m_1$ and $m_2$ on $oPS^\Omega$ by:
\begin{eqnarray}
\label{dJo}
d_{Jo}(m_1,m_2)=\sqrt{\frac{1}{2} (m_1-m_2)^T\underline{\underline{D_o}}(m_1-m_2)},
\end{eqnarray}
where $\underline{\underline{D_o}}$ is given by Equation~\eqref{Do}. 

If we consider $\Omega=\{\omega_1,\omega_2,\omega_3\}$ and the distance given by Equation~\eqref{dMean}, we obtain:
\begin{eqnarray}
 \underline{\underline{D_o}}=\left(
 \begin{array}{ccccccc}
  1 & 0 & 0 & 0 & 0 & 0 & 0 \\
  0 & 1 & \frac{1}{6} & \frac{1}{2} & 0 & \frac{1}{2} & \frac{1}{3} \\
  0 & \frac{1}{6} & 1 & \frac{1}{2} & \frac{1}{6} & \frac{1}{2} & \frac{1}{3} \\
  0 & \frac{1}{2} & \frac{1}{2} & 1 & \frac{1}{12} & \frac{1}{3} & \frac{2}{3} \\
  0 & 0 & \frac{1}{6} & \frac{1}{12} & 1 & \frac{1}{2} & \frac{1}{3} \\
  0 & \frac{1}{12} & \frac{1}{2} & \frac{1}{3} & \frac{1}{2} & 1 & \frac{2}{3} \\
  0 & \frac{1}{3} & \frac{1}{3} & \frac{2}{3} & \frac{1}{3} & \frac{2}{3} & 1\\
  \end{array}
\right)
\end{eqnarray}
Hence:
\begin{equation}
 d_{Jo}(m_{\omega_1},m_{\omega_2})=\sqrt{\displaystyle \frac{5}{6}}\simeq 0.91  \, \mbox{ and } \, d_{Jo}(m_{\omega_1},m_{\omega_3})=1
\end{equation}
In this way, we show that we can take into account the proximity of ordered elements in the distance. 

The modification of the Jaccard dissimilarity, could induce an interpretation of the non-exclusivity of the elements. We will discuss this in section~\ref{fuzzyOrdered}.

\section{Decision and conflict on ordered elements}
\label{decision}
After combining the mass functions, we generally want to make a decision about the resulting mass function. This mass function with the operators seen in section~\ref{combination} takes its values on $oPS^\Omega$. It is common to make the decision on $\Omega$ by maximum of credibility, plausibility or by compromise with the pignistic probability. We have seen in section~\ref{oPS}, that we can calculate these belief functions in $oPS^\Omega$ and if we denote $f_d$ one of these functions the decision is made by:
\begin{eqnarray}
	\omega_d=\argmax_{\omega \in \Omega} \left(f_d(\omega)\right).
\end{eqnarray}

However, these functions do not consider the difference of proximity of the elements of $\Omega$ and do not allow to decide on the disjunctions, {\em i.e.} on some elements of $oPS^\Omega$. In \cite{Essaid14}, we introduced another decision process based on a distance given by:
\begin{eqnarray}
	A=\argmax_{X \in {\cal D}} \left(d_{Jo}(m,m_X)\right),
\end{eqnarray}
where $m_X$ is the categorical mass function $m(X)=1$, $m$ is the mass function coming from the combination rule and $d_{Jo}$ is the distance introduced in Equation~\eqref{dJo}. The subset ${\cal D} \subseteq oPS^\Omega$ is the set of elements on which we want to decide. 

This last decision process also allows to decide on imprecise elements of the ordered power set $oPS^\Omega$ and to take into account with the distance the difference of proximity of ordered element of $\Omega$. 

Let us take the example of the Paris-Lannion trip, if we get 3 categorical answers from 3 people such as: Person 1 said: 314km, person 2 said: 514km, and person 3 said: 614km. Person 1 disagrees  more with person 2 than with person 3. The distance introduced in section~\ref{distoPS} can measure this difference on the ordered elements of the frame of discernment. Therefore, we can extend here the conflict measure introduced in~\cite{Martin12}, between two mass functions $m_1$ and $m_2$ by:
\begin{equation}
 \Conf(m_1,m_2)=(1-\delta_{inc}(m_1,m_2)) d_{Jo}(m_1,m_2)
\end{equation}
where $d_{Jo}$ is the distance defined by equation~\eqref{dJo} and $\delta_{inc}$ is a degree of inclusion measuring how $m_1$ and $m_2$ and included each other. For more details on possible degree of inclusion see~\cite{Martin12}. To measure the conflict between more than two mass functions, the average of the conflicts two by two can be considered. 

\section{Belief functions on ordered fuzzy elements}
\label{fuzzyOrdered}

Now assume that the answers to the question about the distance between Lannion and Paris are:\\
\null \quad \textit{about 300km, about 400km, about 500km, about 600km}\\
The possible answers are always ordered, and if they can always be considered exclusive, they are fuzzy answers. 

The link between belief functions and fuzzy sets has already been discussed in different ways \cite{Denoeux00,Denoeux21}. Here we wish to study the representation and the consideration of ordered and fuzzy elements of the frame of discernment.

As we saw in section~\ref{distoPS} by redefining Jaccard's dissimilarity, the exclusivity of the ordered elements can be questioned. The fact of having ordered fuzzy elements does not however call into question the exclusivity of the elements and thus the intersection of the elements. Indeed, there is no semantic sense in having about 400km and about 500km  at the same time. On the other hand, considering fuzzy elements allows to redefine the cardinality of the intersection of the elements from the membership functions. 

In \cite{Petkovic21}, the cardinality of the intersection of ordered elements is defined from a definition of the membership function, which allows them to propose another modified version of Jaccard dissimilarity. Inspired by this work, we define here the membership function of an element of $\omega\in \Omega$ to a set of $X \in oPS^\Omega$ by:
\begin{equation}
 \mu_X(\omega)=\left\{
 \begin{array}{l}
  1, \, \mbox{if } \, \omega \in X,\\
\\
\displaystyle \alpha e^{- \gamma d_{o}(\omega,X)}, \, \mbox{otherwise},\\
 \end{array}
\right.
\end{equation}
with $0 \leq \alpha \leq 1$, $0 \leq \gamma \leq 1$ two thresholds to control the membership function and $d_{o}$ the distance defined by Equations~\eqref{d_min}, \eqref{d_max} or \eqref{d_mean}. Based on this membership function, we define the cardinality of the intersection of two elements $X$ and $Y \in oPS^\Omega$ by:
\begin{equation}
\label{cardo}
 |X\cap Y|_o = \sum_{\omega \in X \Ocup Y, \omega \in \Omega} \min(\mu_X(\omega),\mu_Y(\omega))
\end{equation}
If $\alpha=0$ then $|X\cap Y|_o =|X\cap Y|$.

Therefore, we define a modified Jaccard dissimilarity for ordered elements by:
\begin{eqnarray}
\label{Dof}
D_o(A,B)=\left\{
\begin{array}{l}
1, \, \mbox{if} \, A= B=\emptyset,\\
\\
\displaystyle \frac{|A\cap B|_o}{|A \Ocup B|}, \, \forall A, B \in oPS^\Omega.\\
\end{array}
\right.
\end{eqnarray}
We have $0 \leq D_o(A,B) \leq 1$ and $D(A,B) \leq D_o(A,B)$~\cite{Petkovic21}. 
A new distance $d_{Jo}$ between the ordered and fuzzy elements can thus be defined by Equation~\eqref{dJo}. 

The cardinality of the intersection defined by Equation~\eqref{cardo} allows its use in the mixed rules~\cite{Martin07} to regulate the conjunctive/disjunctive behaviour by taking into account the partial combinations according to the cardinality of the elements. The mixed rule is given for $m_1$ and $m_2$ for all $X \in 2^\Omega$ by:
\begin{eqnarray}
\label{mix}
\begin{array}{rcl}
m_\mix(X)&=&\displaystyle \sum_{Y_1 \Ocup Y_2 =X} \delta_1 m_1(Y_1)m_2(Y_2)\\
&+& \displaystyle \sum_{Y_1\cap Y_2 =X} \delta_2 m_1(Y_1)m_2(Y_2).
\end{array}
\end{eqnarray}
The choice of $\delta_1=1- \delta_2$ can also be made from Jaccard dissimilarity:
\begin{eqnarray}
\label{eq_d2}
\delta_2(Y_1,Y_2)=\displaystyle \frac{|Y_1\cap Y_2|_o}{|Y_1\Ocup Y_2|}.
\end{eqnarray}
Thus, if we have a partial conflict between $Y_1$ and $Y_2$, $Y_1\cap Y_2=\emptyset$, the rule transfers the mass on $Y_1\Ocup Y_2$ according to the difference in order of the elements of $\Omega$ constituting $Y_1$ and $Y_2$. 

Questionnaires whose responses are in the form of a Likert scale can also be modeled by a frame of discernment of ordered and fuzzy elements and thus use the modified Jaccard dissimilarity defined by Equation~\eqref{Dof}. 

\section{Conclusion}
In this paper, we explored the modeling and integration of an ordered element of frame of discernment by belief functions. The fact of considering ordered elements led us to redefine the power space of unions of ordered elements making sense, called ordered power space. Thus, to consider disjunctive combination rules leaving invariant the ordered power space, we defined the union of ordered elements. From a distance between ordered elements we redefined a distance between mass functions that can be used in a conflict measure and for decision making. Without redefining the intersection of ordered elements, which we still consider exclusive, we have redefined, from a membership function, the cardinality of the intersection of ordered elements that can be considered fuzzy. 

A large number of applications can be addressed by modeling on ordered sets of elements. We are thinking in particular of the answers to questionnaires which very often use Likert scales. This type of questionnaire can, for example, be used to evaluate the knowledge and skills of students allowing to consider the closest answers in the sense of the order of the right answer, without considering them as totally wrong. Concrete applications will be considered in future work.

%
%
%
%

\end{document}